\documentclass{article}
\usepackage{iclr2027_conference,times}
\usepackage{xcolor}
\usepackage{iftex}
\usepackage{fontawesome5}
\ifXeTeX
    \usepackage{xeCJK}

    \setCJKsansfont{FandolHei-Regular.otf}
    \setCJKmonofont{FandolFang-Regular.otf}
\else
    \PackageError{vericua}{This draft must be compiled with XeLaTeX}
      {Select XeLaTeX in the Overleaf compiler settings.}
\fi
\usepackage{enumitem}
\usepackage{amsmath}
\usepackage{amssymb}
\usepackage{booktabs}
\usepackage{float}
\usepackage{fvextra}
\usepackage{graphicx}
\usepackage{microtype}
\usepackage{needspace}
\usepackage{placeins}
\usepackage{colortbl}
\usepackage{tabularx}
\usepackage{widetable}
\usepackage{xurl}
\usepackage{xspace}
\usepackage{hyperref}
\usepackage[capitalise,noabbrev,nameinlink]{cleveref}
\crefname{appendix}{Appendix}{Appendices}
\Crefname{appendix}{Appendix}{Appendices}
\usepackage{wrapfig}

\newcommand{\method}{\textsc{VeriCUA\xspace}}
\newcommand{\judge}{\textsc{Judge}\xspace}
\newcommand{\probe}{\textsc{Probe}\xspace}

\usepackage[breakable,skins]{tcolorbox}

\definecolor{RAPromptText}{HTML}{3A3F45}
\definecolor{RAJudge}{HTML}{3E7F5F}
\definecolor{RAJudgeBg}{HTML}{EDF6F1}
\definecolor{RAProbe}{HTML}{537F9F}
\definecolor{RAProbeBg}{HTML}{EEF4F8}
\definecolor{RARef}{HTML}{9B783D}
\definecolor{RARefBg}{HTML}{FBF7EC}

\DefineVerbatimEnvironment{PromptBlock}{Verbatim}{
    fontsize=\normalsize,
    formatcom=\color{RAPromptText},
    breaklines=true,
    breakanywhere=true,
    breaksymbolleft={},
    baselinestretch=1.05
}

\tcbset{raprompt/.style={
    enhanced, breakable,
    boxrule=0.55pt, arc=1.6mm,
    left=2.4mm, right=2.4mm, top=1.2mm, bottom=1.4mm,
    toptitle=0.9mm, bottomtitle=0.9mm, lefttitle=2.4mm,
    fonttitle=\normalsize\bfseries\sffamily,
    coltitle=white,
    titlerule=0pt,
    before skip=10pt plus 2pt, after skip=10pt plus 2pt,
    pad at break*=1.6mm,
    drop fuzzy shadow=black!12!white
}}
\newtcolorbox{JudgePrompt}[1]{raprompt,
    colframe=RAJudge, colbacktitle=RAJudge, colback=RAJudgeBg,
    title={JUDGE\;\textcolor{white!75!RAJudge}{$\vert$}\; #1}}
\newtcolorbox{ProbePrompt}[1]{raprompt,
    colframe=RAProbe, colbacktitle=RAProbe, colback=RAProbeBg,
    title={PROBE\;\textcolor{white!75!RAProbe}{$\vert$}\; #1}}
\newtcolorbox{RefPrompt}[1]{raprompt,
    colframe=RARef, colbacktitle=RARef, colback=RARefBg,
    title={REFERENCE\;\textcolor{white!75!RARef}{$\vert$}\; #1}}

\definecolor{RAOurs}{HTML}{DCEFE6}
\definecolor{RABackbone}{HTML}{F0F1F3}
\definecolor{RAGain}{HTML}{4C9470}

\title{VeriCUA: Learning Proactive Verification for\\
Reliable GUI Agent Rewards}

\author{Zhizhou Wang$^{1}$\thanks{Equal contribution (\texttt{jenojeff66@gmail.com}, \texttt{lyx23@mails.tsinghua.edu.cn}).}\;,\;
Yaxi Lu$^{2}$\footnotemark[1]\;,\;
Zhong Zhang$^{3}$\thanks{Corresponding author (\texttt{zhong.zhang@uestc.edu.cn}, \texttt{fangminghk@gmail.com}).}\;,\;
Zikun Hu$^{4}$,\;
\textbf{Xin Cong}$^{2}$,\;
\textbf{Yankai Lin}$^{5}$\\
\textbf{Fangming Liu}$^{6}$\footnotemark[2]\\
$^{1}$Southwest Jiaotong University\quad
$^{2}$Tsinghua University\\
$^{3}$University of Electronic Science and Technology of China\\
$^{4}$National University of Singapore\quad
$^{5}$Renmin University of China\quad
$^{6}$Peng Cheng Laboratory}

\iclrfinalcopy

\begin{document}

\maketitle
\lhead{Preprint}

\begin{abstract}
Reinforcement learning for computer-use agents increasingly relies on proactive reward agents that interact with the post-execution GUI environment to verify what recorded screenshots cannot reveal. The reliability of such verification hinges on two coupled decisions: what to check and when to stop. Existing methods leave these decisions to prompted foundation models. However, neither scaling up the model nor providing few-shot examples reliably improves verification: prompted Judges ranging from 4B to GPT-5.4 all plateau at $68\%$--$70\%$ macro-average agreement. This is because the value of a check depends on the environment rather than the task alone, and becomes apparent only through execution. Learning from execution is itself constrained, as the Actor's terminal state cannot be faithfully replicated during training, allowing only one verification attempt per state. We introduce \method{}, a proactive reward agent that learns its verification strategy from these single attempts. A trainable \judge{} decides what to check and when to stop, while a frozen \probe{} executes the checks in the live environment. Online reinforcement learning optimizes the \judge{} for both verdict correctness and efficient \probe{} use, with returns normalized across terminal states. Trained only on OSWorld and evaluated without further training on MobileWorld and WebArena-Verified, \method{} achieves the highest reward agreement across all three benchmarks and lies on the agreement--cost Pareto frontier among proactive verifiers. It surpasses few-shot GPT-5.4 by $5.36$ points with $47.6\%$--$59.8\%$ fewer \probe{} tokens, and even a 1B \method{} exceeds every prompted Judge. As the reward for Actor RL, it raises task success by $3.21$ points overall and $5.40$ points on held-out multi-application tasks. These results identify the verification strategy, not verifier scale, as the key to reliable GUI rewards. 
Code is available at \url{https://github.com/Jeffjeno/VeriCUA}.
\end{abstract}
\section{Introduction}

Training computer-use agents (CUAs) with reinforcement learning requires reliable rewards for task completion~\citep{lv2026scalecuascalingcomputeruse,lai2025computerrl,wang2026cuagymscalingverifiabletraining}. Rule-based rewards are precise but brittle and difficult to scale~\citep{osworld,webarenaverified,mobileworld}, while passive model-based rewards are more flexible but can only verify what is exposed in the available observations~\citep{Qi2025WebRLTL,digirl,distrl,zerogui}. Decisive evidence, such as the contents of a saved file or other hidden application state, may therefore remain unobserved~\citep{smartsnap}. Proactive verification addresses this limitation by interacting with the post-execution environment to acquire the missing evidence before issuing a verdict~\citep{prore,vagen,ira}. Because such verification is performed for every training rollout, it must be not only accurate but also efficient enough to avoid invoking an expensive model for each reward.

\begin{figure}[htpb!]
    \centering
    \includegraphics[width=\linewidth]{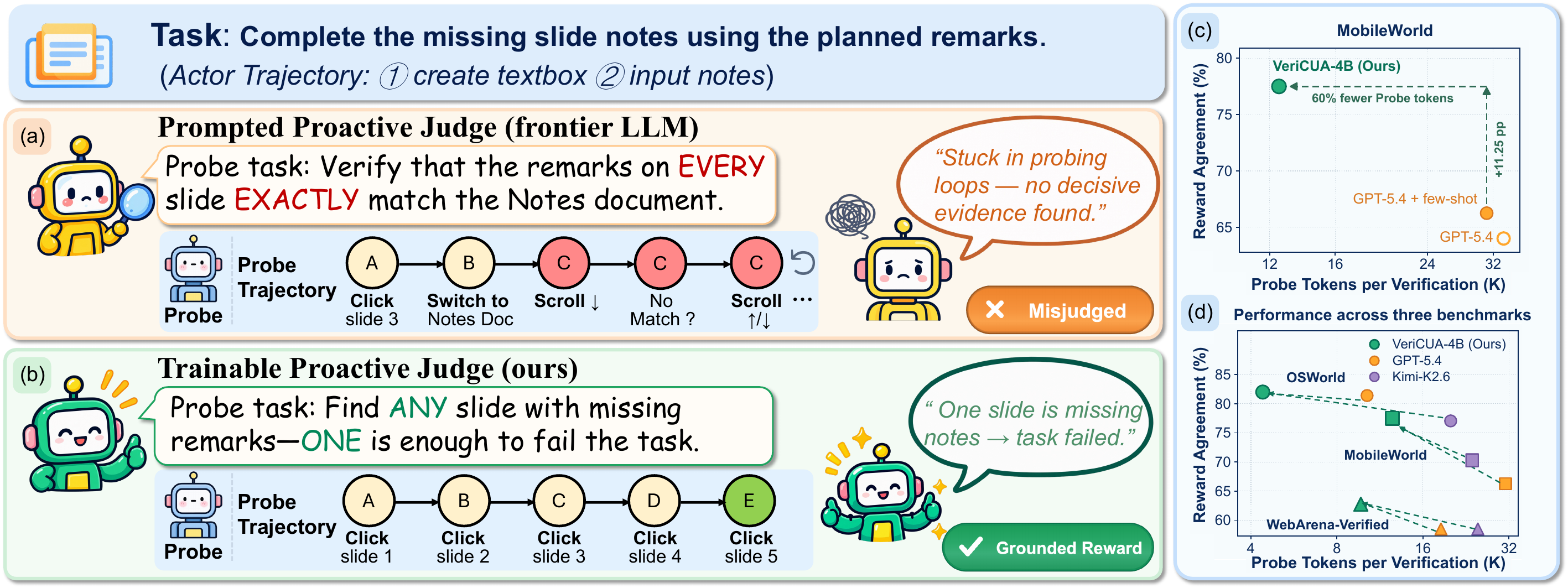}
    \caption{\textbf{Learned check selection and stopping improve verification reliability and efficiency.} Given the same task and frozen \probe{}, (a) prompted GPT-5.4 requests an exhaustive comparison and remains inconclusive, whereas (b) \method{} searches for a slide with missing notes and stops once this counterexample establishes task failure. (c, d) \method{} achieves higher agreement with fewer \probe{} tokens than few-shot GPT-5.4 and Kimi-K2.6.}
    \label{fig:motivation}
\end{figure}

Interaction alone, however, does not guarantee a reliable verdict. Reliability depends on the verification strategy, which consists of two coupled decisions the task leaves open: \textbf{what to check} and \textbf{when to stop}. For a task that requires notes on every slide, one slide without notes already establishes failure (\cref{fig:motivation}(b)). The prompted verifier in \cref{fig:motivation}(a) instead requests a slide-by-slide comparison and keeps scrolling without reaching a verdict. Scale does not fix this. Under the same verification loop, prompted Judges from a 4B model to GPT-5.4 all reach $68\%$--$70\%$ macro-average agreement, and few-shot examples add at most $0.6$ points (\cref{tab:main-consistency}).

Our key observation is that \emph{a check's value is revealed by executing it, not by reading the task}. A prompted Judge sees the task and the recorded screenshots, which show what a plausible check looks like but not what evidence it will return. That depends on where the evidence resides, the state the Actor left, and what the \probe{} can reach. When to stop depends on how often screenshot-only verdicts turn out to be correct. The strategy must therefore be learned by executing verification and comparing its verdicts with true outcomes.

Learning from execution, however, faces a \emph{single-attempt constraint}: each Actor terminal state admits only one verification attempt, which rules out within-state rollout comparisons. An Actor starts from a resettable initial state, so RL can compare several rollouts from the same state~\citep{deepseekmath}. A GUI verifier starts from the Actor's terminal state, the product of one stochastic execution. Re-execution may not reproduce it, identical screenshots can hide different application state, and probing itself changes it.

We introduce \method{}, a proactive GUI reward agent that learns its verification strategy under this constraint. A trainable \judge{} decides what to check and when to stop, and a frozen GUI \probe{} executes the checks. Freezing the \probe{} ties outcome differences to the \judge{} and lets the \probe{} be swapped without retraining. We initialize the \judge{} with outcome-filtered teacher traces and then train it with online RL on complete attempts graded by executable outcomes, so each check is credited by the evidence it actually yields. The reward also depends on whether the \probe{} was used, which lets a single attempt signal whether stopping was right. Returns are normalized across terminal states in each batch~\citep{reinforcepp}, so no environment forking is required.

We evaluate \method{} on OSWorld~\citep{osworld}, MobileWorld~\citep{mobileworld}, and WebArena-Verified~\citep{webarenaverified}. \textbf{First, strategy matters more than scale.} \method{} surpasses few-shot GPT-5.4 by $5.36$ macro-average points with $47.6\%$--$59.8\%$ fewer \probe{} tokens and lies on the agreement--cost Pareto frontier (\cref{fig:pareto}). Even \method{}-1B outperforms prompted GPT-5.4~\citep{openai2026gpt54} and its teacher Kimi-K2.6~\citep{kimi2026k26}. \textbf{Second, the learned strategy transfers.} Trained only on OSWorld, the \judge{} achieves the highest agreement on unseen mobile and web environments. \textbf{Third, better rewards train better agents.} \method{} rewards raise downstream Pass@4 by $3.21$ points overall and by $5.40$ points on held-out MultiApp tasks.

\section{Method}
\label{sec:method}
\subsection{Overview}

\begin{figure}[t]
    \centering
    \includegraphics[width=\linewidth]{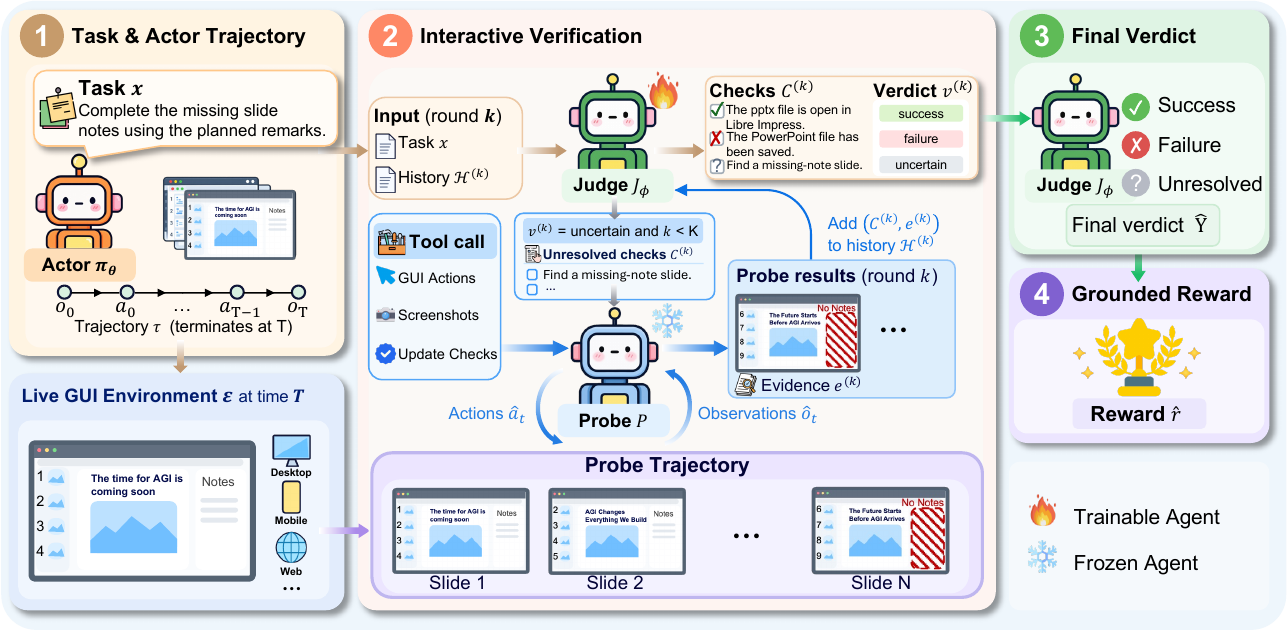}
    \caption{Overview of \method{}. The \judge{} formulates task-success checks and assesses whether the available evidence supports a terminal verdict. When further inspection is requested, the frozen \probe{} gathers evidence for unresolved checks in the live environment and returns status updates with supporting observations.}
    \label{fig:lantern-overview}
\end{figure}

As shown in \cref{fig:lantern-overview}, \method{} verifies the Actor's task outcome using its execution history and live terminal environment. It consists of a trainable \judge{} and a frozen GUI \probe{}. The \judge{} initially formulates task-success checks and marks each as satisfied, violated, or unresolved based on the available evidence. If the evidence supports a success or failure verdict, verification terminates. Otherwise, subject to the probing budget, the \probe{} gathers evidence for the unresolved checks and returns status updates with supporting observations. 
Subsequent \judge{} calls retain the checks from the previous round and update their statuses using the returned evidence.

By specifying what the \probe{} inspects and deciding whether to request further interaction, the \judge{} determines the evidence acquired and hence the verdict. We train it on complete verification attempts, one per Actor terminal state, with a reward on verdict correctness and \probe{} use (\cref{sec:training-recipe,sec:online-judge-rl}).

\subsection{The \texorpdfstring{\judge{}}{Judge}-\texorpdfstring{\probe{}}{Probe} Verification Loop}
\label{sec:interactive-verification}

Consider an Actor policy $\pi_\theta$ attempting task $x$ in a live GUI environment $\mathcal{E}$ and terminating at time $T$ with trajectory $\tau=(o_0,a_0,\ldots,a_{T-1},o_T)$, where $o_t$ and $a_t$ denote GUI observations and actions. The \judge{} receives $x$ and an input representation $\psi(\tau)$ containing the Actor's action history and its most recent screenshots (\cref{sec:prompt-templates}). When the \judge{} requests additional evidence, the frozen \probe{} interacts with the live terminal environment to inspect unresolved task conditions and returns status updates with supporting observations. Each round therefore involves the two decisions that make up the verification strategy: whether the current evidence supports a terminal verdict and, if not, which checks to send to the \probe{}.

\paragraph{\judge{}.}
Let $\mathcal{H}^{(k)}$ denote the verification history available to the \judge{} at call $k$, consisting of the initial Actor input and the checks and \probe{} evidence retained from earlier rounds. At call $k\in\{0,\ldots,K\}$, where $K$ is the maximum number of \probe{} invocations, the trainable \judge{} $J_\phi$ conditions on the task $x$ and $\mathcal{H}^{(k)}$ to produce a verdict and status-annotated checks:
\begin{equation}
    \bigl(v^{(k)},C^{(k)}\bigr)
    = J_\phi\!\left(x,\mathcal{H}^{(k)}\right),
    \qquad
    \mathcal{H}^{(0)}=\psi(\tau).
    \label{eq:judge-verification}
\end{equation}
Here $v^{(k)}\in\{\texttt{success},\texttt{failure},\texttt{uncertain}\}$, and $C^{(k)}$ pairs each check with a status: satisfied, violated, or unresolved. The check text is generated at $k=0$ and the statuses are updated thereafter. The \judge{} returns \texttt{success} when all checks are satisfied, \texttt{failure} when any is violated, and \texttt{uncertain} otherwise. A \texttt{success} or \texttt{failure} verdict terminates verification. When $v^{(k)}=\texttt{uncertain}$ and $k<K$, only the unresolved subset $C_u^{(k)}\subseteq C^{(k)}$ is passed to the \probe{}.

\paragraph{\probe{}.}
The frozen \probe{} $P$ gathers evidence for the unresolved checks in the current live environment. We keep it frozen so that differences in verification outcome are attributable to the \judge{}. Its instructions confine it to inspecting the environment, without completing or repairing the Actor's task (\cref{sec:prompt-templates}). Its interaction trajectory and returned evidence are represented as
\begin{equation}
    \bigl(\tau_P^{(k)},e^{(k)}\bigr)
    \sim
    P\!\left(\cdot\mid C_u^{(k)};\mathcal{E}^{(k)}\right),
    \qquad
    \mathcal{H}^{(k+1)}
    =
    \mathcal{H}^{(k)}\oplus\bigl(C^{(k)},e^{(k)}\bigr).
    \label{eq:probe-evidence}
\end{equation}
Here $\mathcal{E}^{(0)}$ is the Actor's terminal environment, and each subsequent \probe{} invocation continues from the state reached by the preceding \probe{} actions. The returned evidence $e^{(k)}$ comprises check-status updates and selected supporting observations, and $\oplus$ denotes concatenation. Only $e^{(k)}$ reaches the \judge{}; the full \probe{} trace $\tau_P^{(k)}$ does not, and each invocation starts without the interaction history of earlier rounds, so neither component's context grows with the other's execution length.

\subsection{Outcome Feedback from a Single Attempt}
\label{sec:training-recipe}

We assign a terminal reward to each complete verification attempt, combining verdict correctness with whether probing was used. A correctness-only reward scores a correct verdict reached directly the same as one reached after probing. Because each terminal state admits a single attempt, there is no sibling attempt to reveal whether probing was needed, so the reward itself must carry that signal.

Let $\hat Y_i$ denote the final \judge{} verdict and $Y_i$ the executable outcome recorded before \probe{} interaction. We set $p_i=1$ if the verification attempt invokes the \probe{} at least once, and $p_i=0$ otherwise, so the probing term of the reward reflects the \judge{}'s stopping decision rather than the frozen \probe{}'s execution length. The terminal reward is $r_i=R(\hat Y_i,Y_i,p_i)$.

Correct verdicts receive higher rewards when reached without probing, encouraging reliable judgments from existing evidence. Incorrect verdicts receive larger penalties when issued without probing, discouraging premature judgments. Unresolved or invalid outputs receive the lowest reward, discouraging attempts that exhaust the probing budget without reaching a terminal verdict. The exact reward schedule is given in \cref{sec:verification-reward-details}.

We use this terminal reward to jointly train check formulation and stop-or-probe decisions over verification attempts. The reward preferences teach the \judge{} to stop once the evidence suffices, while the shared return trains check formulation through its effect on the evidence collected and the verdict. We next describe how we collect verification trajectories and optimize the \judge{}.

\begin{figure}[t]
    \centering
    \includegraphics[width=\linewidth]{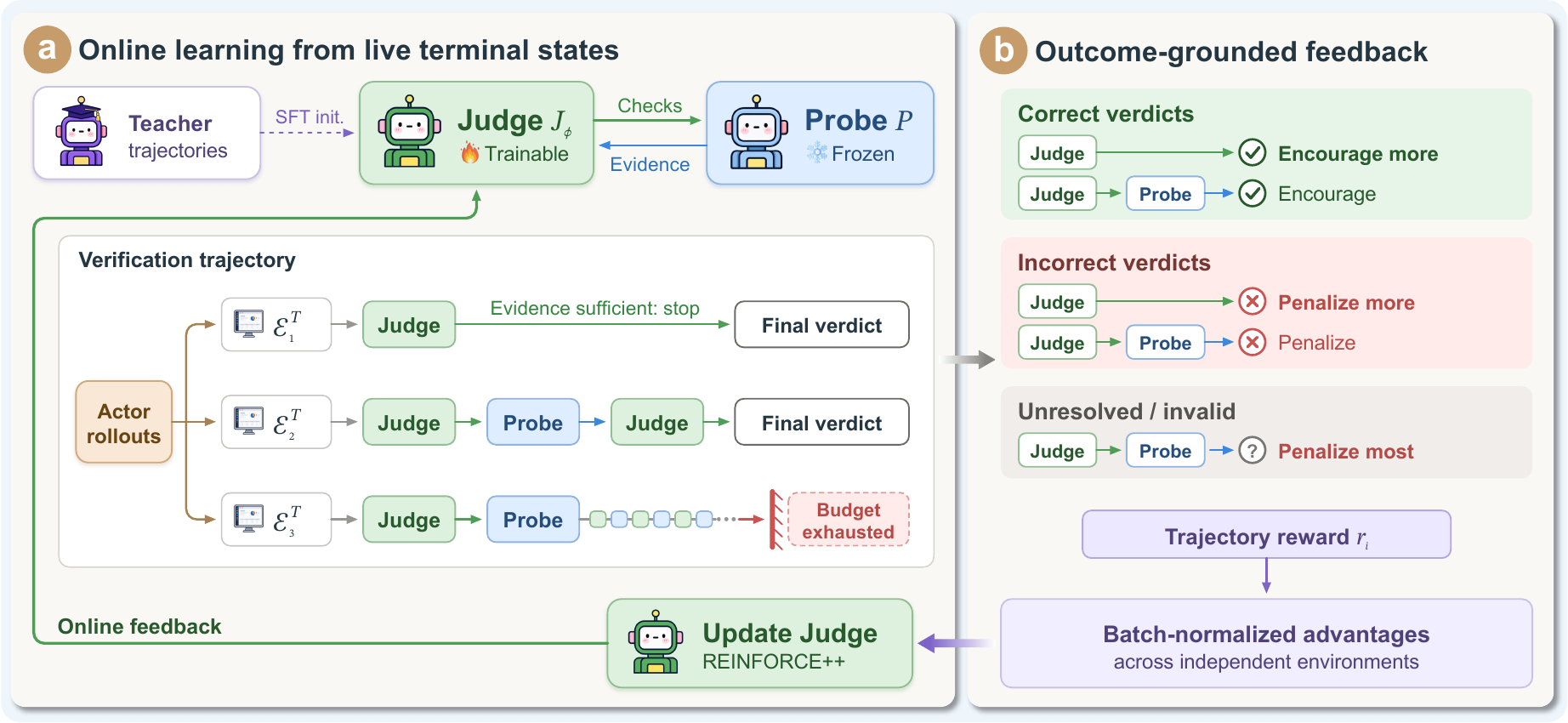}
    \caption{Training procedure of \method{}. Online optimization uses one verification trajectory per live Actor terminal state. The \judge{} is updated using terminal feedback on verdict correctness and whether probing was used, while the \probe{} remains frozen.}
    \label{fig:training-recipe}
\end{figure}

\subsection{Online Optimization of the \texorpdfstring{\judge{}}{Judge}}
\label{sec:online-judge-rl}

\paragraph{Initialization.}
\label{sec:sft-coldstart}
Before online optimization, we initialize the \judge{} by supervised fine-tuning on teacher verification trajectories whose final verdicts match the executable outcomes. These traces provide examples of structured check generation and verdict prediction, so that each attempt during online RL is a complete, well-formed verification. The \probe{} remains frozen.

\paragraph{Sampling verification trajectories.}
We run the \judge{} from the live terminal environment of each Actor rollout and execute its \probe{} requests to obtain the evidence. Each completed verification attempt receives the terminal reward defined in \cref{sec:training-recipe}. Unlike an Actor's resettable initial state, this terminal environment is produced by one stochastic execution: replaying the Actor need not reproduce it, transient state such as unsaved edits and modal dialogs cannot be confirmed from screenshots, and \probe{} actions further alter it. We therefore collect one verification trajectory per Actor terminal environment, without environment forking or same-state verifier rollout groups.

\paragraph{Updating the \judge{}.}
We optimize the \judge{} with REINFORCE++~\citep{reinforcepp}, which normalizes terminal returns across the independent terminal environments in each batch. Let $z_i=\bigl((v_i^{(0)},C_i^{(0)}),e_i^{(0)},\ldots,(v_i^{(K_i)},C_i^{(K_i)})\bigr)$ denote a complete verification trajectory with $K_i\leq K$ \probe{} invocations and terminal reward $r_i$. When $K_i=0$, the trajectory contains only the initial \judge{} output. For a batch $\mathcal{B}$ of complete verification trajectories, the normalized return supplies the advantage for every active \judge{} token:
\begin{equation}
    A_{i,k,t}
    =
    \frac{r_i-\mu_{\mathcal{B}}}
         {\sqrt{\sigma_{\mathcal{B}}^2+\epsilon}},
    \qquad
    (i,k,t)\in\mathcal{A}_{\mathcal{B}}.
    \label{eq:judge-whitening}
\end{equation}
Here $\mu_{\mathcal{B}}$ and $\sigma_{\mathcal{B}}^2$ are the mean and variance of terminal rewards in the batch, and $\epsilon>0$ ensures numerical stability. The index set $\mathcal{A}_{\mathcal{B}}$ contains active \judge{} tokens, indexed by trajectory $i$, \judge{} call $k$, and token position $t$. Only newly generated \judge{} tokens contribute to the loss; context tokens and \probe{} outputs are excluded. The following log-probability surrogate expresses the on-policy gradient at the rollout policy:
\begin{equation}
    \mathcal{L}_{\mathrm{Judge}}(\phi)
    =
    -\frac{1}{|\mathcal{A}_{\mathcal{B}}|}
    \sum_{(i,k,t)\in\mathcal{A}_{\mathcal{B}}}
    A_{i,k,t}
    \log \pi_{\phi}^{J}
    \left(y_{i,k,t}\mid h_{i,k,t}\right),
    \label{eq:judge-policy-gradient}
\end{equation}
where $\pi_{\phi}^{J}$ is the autoregressive \judge{} policy, $y_{i,k,t}$ is the generated token, and $h_{i,k,t}$ is its conditioning context. The loss averages over active \judge{} tokens across the batch. For parameter updates, we use the clipped likelihood-ratio objective described in \cref{sec:reinforcepp-details}, with rollout-policy probabilities and normalized returns held fixed.

\paragraph{Using the trained verifier.}
\label{sec:downstream-policy-learning}
After training, we freeze \method{} and map its final verdict to a binary reward
$\hat r_i = \mathbb{I}\!\left[\method{}(x_i,\tau_i,\mathcal{E}_i^T) = \texttt{success}\right]$,
where $\mathcal{E}_i^T$ is the terminal environment of Actor rollout $i$.
We keep \method{} fixed and optimize the Actor with vanilla
GRPO~\citep{deepseekmath}, grouping Actor rollouts generated from the
same task initialization. Each rollout is verified once from its own
terminal environment, and the resulting verdict determines its reward $\hat r_i$.

\section{Experiments}
\label{sec:experiments}

Our experiments examine whether a learned verification strategy improves reward agreement over prompted verifiers across model scales and at what \probe{} cost, whether it generalizes beyond its training environment, and whether the resulting rewards improve downstream Actor learning.

\subsection{Experimental Setup}
\label{sec:experimental-setup}

\begin{table}[t]
    \centering
    \caption{Reward reliability across executable environments (\%). Results use Avg@$4$ except ZeroGUI$^{*}$, which follows its official Majority@$4$ aggregation. \method{} leads on all three benchmarks. Bold and underlined mark the best and second-best results.}
    \label{tab:main-consistency}
    \small
    \setlength{\tabcolsep}{1.7pt}
    \renewcommand{\arraystretch}{0.95}
    \begin{tabularx}{\linewidth}{@{}>{\raggedright\arraybackslash}p{0.25\linewidth}>{\raggedright\arraybackslash}p{0.105\linewidth}*{4}{>{\centering\arraybackslash}X}@{}}
        \toprule
        \textbf{Verifier} & \textbf{Mode}
        & \begin{tabular}[c]{@{}c@{}}\footnotesize\textbf{OSWorld}\\[-2pt]\scriptsize desktop\end{tabular}
        & \begin{tabular}[c]{@{}c@{}}\footnotesize\textbf{MobileWorld}\\[-2pt]\scriptsize mobile\end{tabular}
        & \begin{tabular}[c]{@{}c@{}}\footnotesize\textbf{WebArena-V}\\[-2pt]\scriptsize web\end{tabular}
        & \begin{tabular}[c]{@{}c@{}}\footnotesize\textbf{Macro Avg.}\\[-2pt]\scriptsize all envs.\end{tabular} \\
        \midrule
        \rowcolor{RABackbone}
        \multicolumn{2}{@{}l}{\textit{GPT-5.4}} & \multicolumn{4}{r@{}}{\scriptsize proprietary reference} \\
        Prompted \method{} & Prompted & \underline{81.88} & 64.00 & 58.50 & 68.13 \\
        \quad + few-shot & Prompted & 81.40 & 66.25 & 58.50 & 68.72 \\
        \midrule
        \rowcolor{RABackbone}
        \multicolumn{2}{@{}l}{\textit{Kimi-K2.6}} & \multicolumn{4}{r@{}}{\scriptsize proprietary reference} \\
        Prompted \method{} & Prompted & 75.85 & 74.00 & 59.00 & 69.62 \\
        \quad + few-shot & Prompted & 77.05 & 70.25 & 58.50 & 68.60 \\
        \midrule
        \rowcolor{RABackbone}
        \multicolumn{2}{@{}l}{\textit{Qwen3-VL-4B-Thinking}} & \multicolumn{4}{r@{}}{\scriptsize 4B main backbone} \\
        DistRL & Passive & 61.78 & 68.00 & 57.75 & 62.51 \\
        DigiRL & Passive & 63.71 & 68.25 & 57.75 & 63.24 \\
        ZeroGUI$^{*}$ & Passive & 71.50 & 73.00 & 51.00 & 65.17 \\
        VAGEN & Prompted & 65.34 & 70.25 & 56.25 & 63.95 \\
        Prompted \method{} & Prompted & 73.73 & 73.50 & 61.50 & 69.58 \\
        \rowcolor{RAOurs}
        \method{} (ours) & Trained & \textbf{82.00} & \textbf{77.50} & \textbf{62.75} & \textbf{74.08} \\
        \rowcolor{RAOurs}
        \multicolumn{2}{@{}l}{\textcolor{RAGain}{\footnotesize$\Delta$ vs. Prompted (pp)}} & \textcolor{RAGain}{\footnotesize+8.27} & \textcolor{RAGain}{\footnotesize+4.00} & \textcolor{RAGain}{\footnotesize+1.25} & \textcolor{RAGain}{\footnotesize+4.50} \\
        \midrule
        \rowcolor{RABackbone}
        \multicolumn{2}{@{}l}{\textit{Qwen3-VL-2B-Thinking}} & \multicolumn{4}{r@{}}{\scriptsize 2B main backbone} \\
        Prompted \method{} & Prompted & 66.49 & 65.00 & 55.00 & 62.16 \\
        \rowcolor{RAOurs}
        \method{} (ours) & Trained & 77.84 & 72.50 & 62.00 & 70.78 \\
        \rowcolor{RAOurs}
        \multicolumn{2}{@{}l}{\textcolor{RAGain}{\footnotesize$\Delta$ vs. Prompted (pp)}} & \textcolor{RAGain}{\footnotesize+11.35} & \textcolor{RAGain}{\footnotesize+7.50} & \textcolor{RAGain}{\footnotesize+7.00} & \textcolor{RAGain}{\footnotesize+8.62} \\
        \midrule
        \rowcolor{RABackbone}
        \multicolumn{2}{@{}l}{\textit{MiniCPM-V-4.6-Thinking}} & \multicolumn{4}{r@{}}{\scriptsize 1B main backbone} \\
        Prompted \method{} & Prompted & 52.29 & 58.75 & 40.50 & 50.51 \\
        \rowcolor{RAOurs}
        \method{} (ours) & Trained & 77.11 & \underline{75.50} & \underline{62.25} & \underline{71.62} \\
        \rowcolor{RAOurs}
        \multicolumn{2}{@{}l}{\textcolor{RAGain}{\footnotesize$\Delta$ vs. Prompted (pp)}} & \textcolor{RAGain}{\footnotesize+24.82} & \textcolor{RAGain}{\footnotesize+16.75} & \textcolor{RAGain}{\footnotesize+21.75} & \textcolor{RAGain}{\footnotesize+21.11} \\
        \bottomrule
    \end{tabularx}
\end{table}

\paragraph{Benchmarks.}
We evaluate on the OSWorld~\citep{osworld} tasks that require neither network access nor Google Drive, the GUI-only subset of MobileWorld~\citep{mobileworld}, and WebArena-Verified Hard~\citep{webarenaverified}, using $414$, $100$, and $100$ trajectories, respectively.
These cover successful executions and diverse failures from five Actors: EvoCUA-32B~\citep{xue2026evocua}, Qwen3.8-27B~\citep{qwen38}, Holo3.1-35B-A3B~\citep{hai2026holo31}, Qwen3-VL-4B-Thinking~\citep{Qwen3-VL}, and Gemini 3.7 Flash~\citep{google2026gemini37flash}.
Each trajectory includes the action--observation history, the clean task snapshot from which its terminal environment can be re-executed, and an executable outcome label hidden from all verifiers.

\paragraph{Training data.}
We train the \judge{} only on OSWorld and keep it frozen thereafter, so MobileWorld and WebArena-Verified test zero-shot transfer. Each environment uses one frozen \probe{} shared by all \method{} variants.
We denote trained Judges by size: \method{}-4B (Qwen3-VL-4B-Thinking), \method{}-2B (Qwen3-VL-2B-Thinking), and \method{}-1B (MiniCPM-V-4.6-Thinking~\citep{yu2025minicpmv45cookingefficient,yao2024minicpm}).
SFT uses Kimi-K2.6 verification traces and online RL uses newly collected OSWorld trajectories, all disjoint from evaluation (\cref{sec:lantern-training-details}).

\paragraph{Baselines.}
Unless noted, all baselines use Qwen3-VL-4B-Thinking, and existing methods follow their published observation and prompting configurations (\cref{sec:evaluation-details}).
\textbf{(i) Passive verifiers:} DistRL~\citep{distrl}, DigiRL~\citep{digirl}, and ZeroGUI~\citep{zerogui}.
\textbf{(ii) Prompted proactive verifiers:} VAGEN~\citep{vagen} and prompted \method{}, which shares trained \method{}'s \judge{}--\probe{} architecture, prompts, and budget of at most $K{=}2$ \probe{} rounds but uses an off-the-shelf Judge.
We also evaluate prompted \method{} with the backbones of \method{}-2B and \method{}-1B, and with the frontier models Kimi-K2.6 and GPT-5.4, both also tested few-shot.

\paragraph{Evaluation protocol.}
Both evaluations use the executable task evaluator as ground truth and aggregate four independent runs (\cref{sec:evaluation-metrics}).
\textbf{Verifier evaluation:}
We keep only Actor trajectories whose outcome is stable across six re-executions from the clean task snapshot (\cref{sec:evaluation-details}). Each proactive-verifier trial runs in its own re-executed terminal environment, so no verifier's \probe{} actions affect another's. Passive verifiers score the recorded screenshots.
We report Avg@$4$ agreement per benchmark and its unweighted mean (Macro Avg.), and measure cost as frozen-\probe{} tokens per judgment (\cref{sec:evaluation-details}).
\textbf{Actor evaluation:}
We train the same Qwen3-VL-4B-Thinking Actor on OSWorld with vanilla GRPO~\citep{deepseekmath} under identical tasks, rollout budget, and hyperparameters, varying only the frozen reward source (ZeroGUI, prompted \method{}, or trained \method{}). We report Pass@$4$ on the $206$ single-application tasks used for Actor training (ID) and $74$ MultiApp tasks held out from Actor training (OOD).

\paragraph{Implementation details.}
We implement \judge{} SFT with LLaMA-Factory and online RL with Slime,
using Megatron Core for parameter updates and SGLang/vLLM for rollout generation with NVIDIA A800 GPUs and Huawei Ascend 910B3 NPUs.
Detailed hardware, software, and desktop-environment configurations are
provided in \cref{sec:implementation-details}.

\begin{figure}[t]
    \centering
    \includegraphics[width=\linewidth]{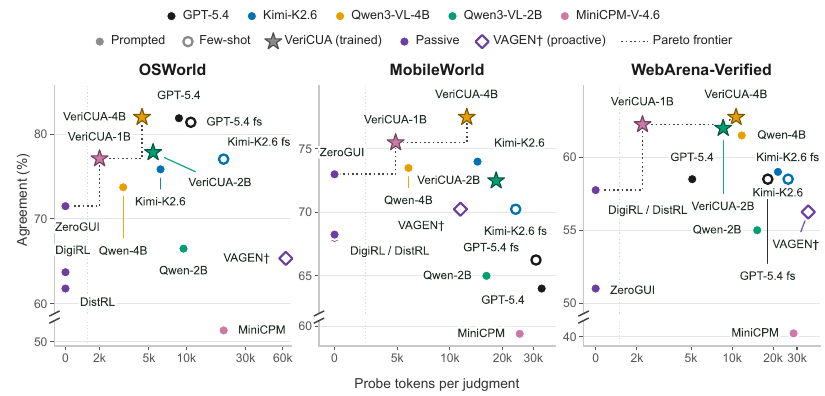}
    \caption{Reward agreement versus \probe{} tokens per judgment: trained \method{} occupies the agreement--cost Pareto frontier among proactive verifiers. VAGEN$^{\dagger}$ is billed for its full loop.}
    \label{fig:pareto}
\end{figure}

\subsection{Verification Strategy Matters More Than Scale}
\label{sec:main-results}

\paragraph{Agreement and \probe{}-token cost.}
Prompted Judges from a 4B model to GPT-5.4 plateau at $68.13\%$--$69.62\%$ macro-average agreement, and few-shot examples shift this range by at most $1.02$ points. \method{}-4B moves beyond this plateau: it achieves the highest agreement on all three benchmarks (\cref{tab:main-consistency}) and lies on the agreement--cost Pareto frontier of proactive verifiers (\cref{fig:pareto}). Compared with few-shot GPT-5.4, it improves macro-average agreement by $5.36$ points ($68.72\%\rightarrow74.08\%$) with $47.6\%$--$59.8\%$ fewer \probe{} tokens. On OSWorld, where GPT-5.4 is strongest, \method{}-4B still leads while using $56.9\%$ fewer tokens, and its margin widens to $11.25$ and $4.25$ points on MobileWorld and WebArena-Verified. Even \method{}-1B ($71.62\%$) exceeds every prompted Judge.

\paragraph{Effect of the learned verification policy.}
With the \judge{} backbone, \probe{}, prompts, and round budget held fixed, training improves agreement on every benchmark. Macro-average agreement rises by $4.50$, $8.62$, and $21.11$ points over the prompted 4B, 2B, and 1B Judges, respectively, so the gains come from the learned verification policy rather than the architecture. The agreement--cost frontier reflects this: on every benchmark, its highest-agreement point is a trained \method{} \judge{}, and \method{}-1B stays on the frontier throughout, dominating every prompted counterpart in both agreement and cost (\cref{fig:pareto}).

\subsection{Generality and Behavior of the Learned Strategy}

\paragraph{Transfer across environments.}
MobileWorld and WebArena-Verified are unseen during training. On both, \method{}-4B and \method{}-1B, paired with environment-specific Probes, rank first and second, ahead of every prompted and passive verifier including few-shot GPT-5.4. The learned strategy is therefore not specific to desktop applications.

\begin{figure}[t]
    \centering
    \includegraphics[width=\linewidth, trim=0 10pt 0 0, clip]{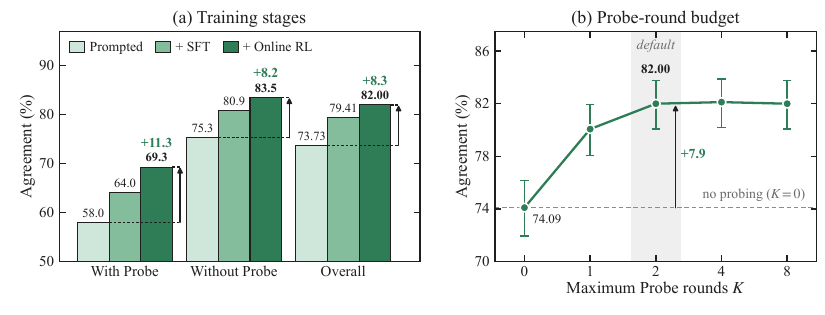}
    \caption{Learning ablations on OSWorld. \textbf{(a)} SFT improves all judgments; online RL mainly lifts those that invoke the \probe{}. \textbf{(b)} Agreement saturates at $K{=}2$, with most gain from the first round.}  
    \label{fig:learning-ablation}
\end{figure}

\paragraph{Contribution of training stages.}
\cref{fig:learning-ablation}(a) separates SFT initialization from online RL for the Qwen3-VL-4B-Thinking Judge on OSWorld. Agreement rises from $73.73\%$ for the prompted Judge to $79.41\%$ after SFT and $82.00\%$ after online RL, so both stages contribute, with online RL adding $2.59$ points. Agreement improves on judgments both with and without probing. Every trained \judge{} also surpasses Kimi-K2.6 ($69.62\%$ macro average), whose traces initialize it, suggesting that one attempt per terminal state suffices to learn beyond imitation.

\paragraph{Effect of the probing budget.}
\cref{fig:learning-ablation}(b) varies the maximum number of \probe{} rounds for the trained 4B Judge. Agreement increases from $74.09\%$ at $K{=}0$ to $82.00\%$ at $K{=}2$, indicating that evidence from the live environment is decisive. The first round accounts for most of the gain, and two rounds suffice. Because $K$ only caps interaction, the \judge{} can still stop earlier.

\paragraph{Robustness to \probe{} size.}
We keep the trained \judge{} fixed and evaluate Holo3.1 Probes at 35B-A3B, 9B, and 4B scales. OSWorld agreement stays between $82.00\%$ and $82.73\%$ across these Probes without retraining (\cref{sec:probe-sensitivity}). This suggests that the learned strategy resides in the \judge{}, while the \probe{} mainly executes it.

\begin{figure}[!htb]
    \centering
    \includegraphics[width=0.95\linewidth, trim=0 10pt 0 0, clip]{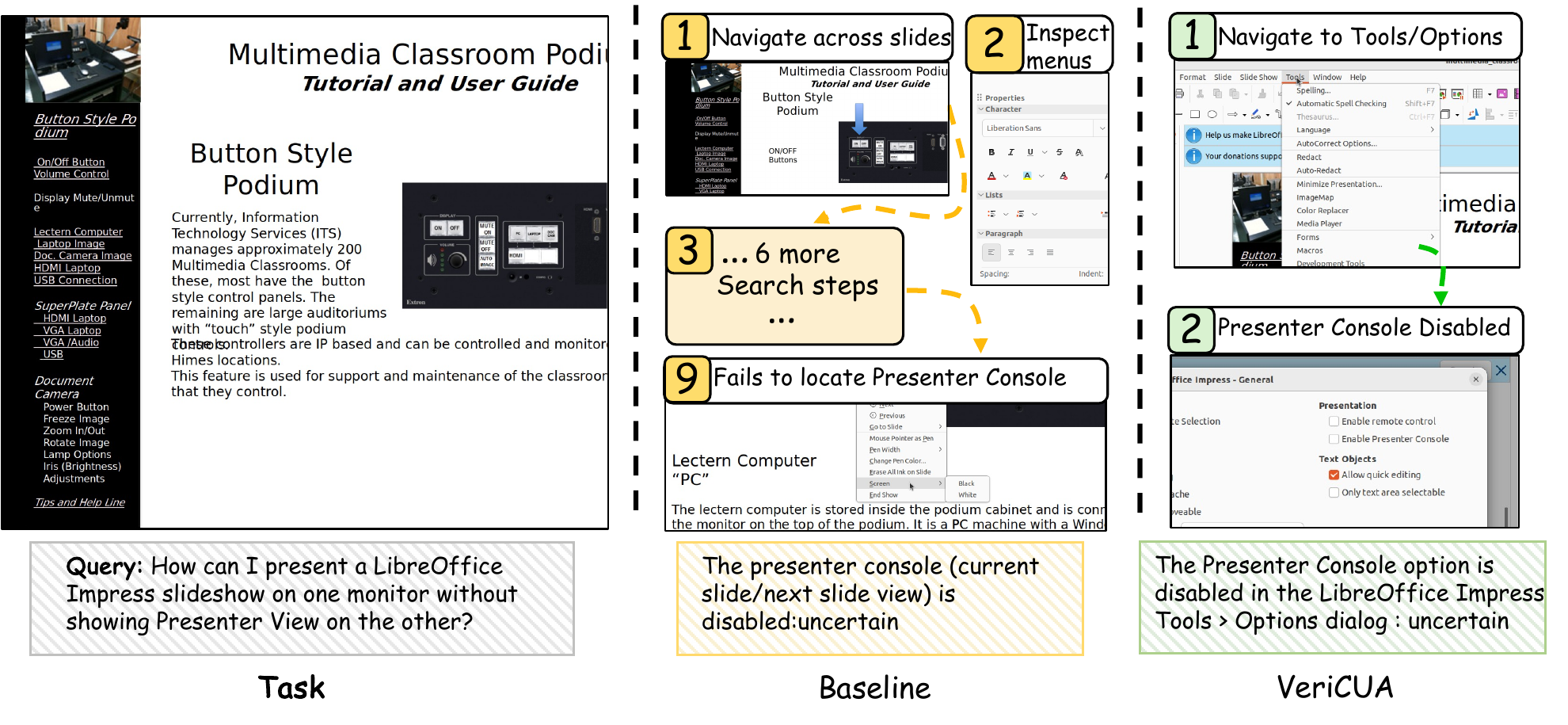}
\caption{Case study. The baseline check specifies only the condition, and the \probe{} fails after nine steps. The \method{} check also specifies the evidence location, enabling verification in two~steps.}    \label{fig:verification-case-study}
\end{figure}

\paragraph{Evidence acquisition behavior.}
We further examine $33$ OSWorld cases in which the recorded Actor trajectory omits outcome-determining evidence that remains accessible through GUI interaction (\cref{sec:verification-behavior}). With the same frozen \probe{}, \method{}-4B directs $61.9\%$ of \probe{} actions toward task-relevant evidence, compared with $47.6\%$ for prompted GPT-5.4. Aimless scrolling and waiting fall from $16.2\%$ to $5.5\%$, with fewer repeated targets and dead-end probes. \cref{fig:verification-case-study} illustrates the difference. Verifying that Presenter Console is disabled requires a setting absent from the recorded trajectory. \method{}'s check names both the condition and its location (\texttt{Tools} $\rightarrow$ \texttt{Options}), and the \probe{} inspects it in two steps. Given the condition without the location, the \probe{} takes nine steps and never finds it.

\subsection{Downstream Actor Learning}
\label{sec:exp-downstream-policy}

\begingroup
\setlength{\columnsep}{10pt}
\setlength{\intextsep}{2pt}
\setlength{\abovecaptionskip}{0pt}
\setlength{\belowcaptionskip}{3pt}
\Needspace{12\baselineskip}

\begin{wraptable}[12]{r}{0.53\textwidth}
\centering
\caption{Pass@$4$ (\%) of the RL-trained actor under different reward sources. $\Delta$ denotes the absolute improvement over the initial actor on the All split.}
\label{tab:downstream-policy}
\setlength{\tabcolsep}{3.5pt}
\definecolor{GainBlue}{HTML}{4F81BD}
\newcommand{\gain}[1]{\textcolor{GainBlue}{$+#1$}}
\begin{tabular}{@{}lcccc@{}}
\toprule
\textbf{Reward source} & \textbf{All} & \textbf{ID} & \textbf{OOD} & $\boldsymbol{\Delta}$ \\
\midrule
None (initial policy)     & 43.93 & 53.40 & 17.57 & -- \\
ZeroGUI                   & 45.36 & 54.37 & 20.27 & \gain{1.43} \\
\midrule
\method{} (prompted)      & 45.71 & 54.85 & 20.27 & \gain{1.78} \\
\method{} (trained) & \textbf{47.14} & \textbf{55.83} & \textbf{22.97} & \textcolor{GainBlue}{$\mathbf{+3.21}$} \\
\bottomrule
\end{tabular}
\end{wraptable}

\paragraph{Actor performance.} Using \method{} rewards yields the highest Pass@$4$ among the compared reward sources on every split (\cref{tab:downstream-policy}). Overall Pass@$4$ rises from $43.93\%$ for the initial Actor to $47.14\%$, a gain of $3.21$ percentage points. On the MultiApp split, it rises from $17.57\%$ to $22.97\%$, a gain of $5.40$ percentage points. The Actor trained with prompted \method{} reaches $45.71\%$ overall and $20.27\%$ on MultiApp, below the corresponding results with the trained verifier.

\endgroup

\paragraph{Verification during Actor training.}
To check whether this advantage persists as the Actor changes, we track verification behavior throughout the runs with prompted and trained \method{} rewards (details in \cref{sec:behavior-analysis}). With the trained verifier, agreement with executable outcomes is higher ($81.4\%$ vs.\ $77.8\%$), the false-negative rate is lower ($4.6\%$ vs.\ $10.0\%$), and the \probe{}-attempt rate is lower ($4.28\%$ vs.\ $5.70\%$). The trained verifier thus probes less often while missing fewer successes, consistent with a reward that scores both verdict correctness and \probe{} use.

\section{Related Work}

\textbf{Reward signals for computer-use agents.}
Task-completion rewards for computer-use agents can be obtained from executable evaluators or model-based judgments. Executable evaluators derive outcome labels from application or backend state~\citep{osworld,mobileworld,webarenaverified}, relying on task-specific checking logic. Environment synthesis automates the construction of verifiable tasks by generating resettable tasks together with executable evaluators~\citep{guigenesis,infiniteweb}. Model-based evaluators assess recorded observations and action histories using prompted or trained judges~\citep{digirl,distrl,Qi2025WebRLTL,zerogui,agentrm}, while rubric and milestone-based critics further structure this assessment~\citep{cuaverifier,osthemis}. However, since recorded observations obscure uncaptured state, verifiers must acquire additional evidence through active interaction with the environment.

\textbf{Interactive verification and evidence acquisition.}
Recent work incorporates evidence acquisition into both post-execution verification and task execution. Post-execution verifiers continue from the Actor's terminal environment and inspect state before issuing a verdict~\citep{prore,vagen,ira}. Other approaches integrate evidence collection, completeness checking, or recovery into the acting policy~\citep{smartsnap,vlaagui}. Text-reasoning verifiers can be trained with RL by replicating a problem to compare verification attempts~\citep{agentvrl}; we study the post-execution GUI setting, where each state yields a single verification attempt.

\section{Conclusion}
We introduced \method{}, a proactive GUI reward agent that learns what to check and when to stop. Prompted verifiers benefit little from larger models or few-shot examples because a check's value is revealed only through environment interaction. With only one verification attempt per Actor terminal state during training, \method{} learns this strategy by comparing outcomes across states. Across desktop, mobile, and web, it reaches the agreement--cost Pareto frontier, enables a 1B Judge to surpass prompted frontier models, and improves downstream policy RL. These results identify verification strategy, rather than model scale, as the key to reliable and affordable GUI reward supervision.

\Needspace{8\baselineskip}
\section*{Reproducibility Statement}
The interactive verification procedure and online training objective are
specified in \cref{sec:interactive-verification,sec:online-judge-rl}.
\Cref{sec:experimental-setup} describes the benchmark splits, baselines,
and re-execution-based evaluation protocol. The appendix provides evaluation
metrics, training and inference settings, the terminal reward schedule,
and complete \judge{} and \probe{} prompts. 

\section*{AI Use Statement}

Large language model (LLM) tools
were used in a supporting role to assist with coding and debugging in the
reinforcement-learning framework, and to polish the manuscript's writing for
clarity and readability. Kimi-K2.6 was also used to generate candidate
verification trajectories for the rejection-sampled SFT cold-start dataset.
LLM tools were not used for brainstorming or generating
research ideas. All AI-assisted textual edits were reviewed by the authors,
and all AI-assisted code changes were reviewed and tested. The authors take
full responsibility for the final content of this paper.

\section*{Acknowledgement}
This work was supported in part by the Natural Science Foundation of China (Grant No. 62606076), in part by the the Major Key Project of PCL under Grant  PCL2025A10, in part by the Shenzhen Science and Technology Program under Grant RCJC20231211085918010, and partially supported by Tsinghua KA Excellence Center.

\bibliographystyle{iclr2027_conference}
\bibliography{references}

\newpage
\appendix
\section{Evaluation Metrics}
\label[appendix]{sec:evaluation-metrics}

For reward evaluation, let $\{(\tau_i,Y_i)\}_{i=1}^{N}$ be a set of Actor trajectories paired with executable evaluator outcomes.  We run each verifier independently $k$ times on every trajectory and denote its $j$-th verdict by $\hat{Y}_{i,j}$.  Each proactive-verifier trial starts from a fresh re-execution of an accepted trajectory (\cref{sec:experimental-setup}). Passive verifiers score the fixed recorded observations.  We report average agreement with the executable evaluator as
\begin{equation}
    \mathrm{Avg@}k
    =
    \frac{100}{Nk}
    \sum_{i=1}^{N}
    \sum_{j=1}^{k}
    \mathbb{I}[\hat{Y}_{i,j}=Y_i].
    \label{eq:avg-at-k}
\end{equation}
Avg@$k$ thus measures how often verifier judgments match the executable outcome, not how often the Actor succeeds.
ZeroGUI follows its official Majority@$4$ aggregation, as marked in
\cref{tab:main-consistency}.

For downstream Actor evaluation, let $\mathcal{D}$ be the task set and let $S_{i,j}\in\{0,1\}$ indicate whether the $j$-th Actor rollout on task $i$ succeeds.  We report the fraction of tasks solved at least once within $k$ rollouts as
\begin{equation}
    \mathrm{Pass@}k
    =
    \frac{100}{|\mathcal{D}|}
    \sum_{i\in\mathcal{D}}
    \mathbb{I}\!\left[\sum_{j=1}^{k}S_{i,j}>0\right].
    \label{eq:pass-at-k}
\end{equation}
Both metrics are reported as percentages, and differences between them in percentage points.

\section{Behavioral Evidence for Targeted Verification}
\label[appendix]{sec:verification-behavior}

We analyze 33 OSWorld cases in which the recorded trajectory leaves the
outcome-determining state unobserved, but the missing evidence can still be
reached through the GUI. A correct verdict in these cases requires the
\judge{} to find an effective verification path and to recognize when the
returned evidence is sufficient.

Reward agreement and the \probe{}-budget ablation measure only the final
verdict. To see how verification interaction is spent along the way, we
measure four behavioral proxies (\cref{tab:verification-proxies}).

\paragraph{What the proxies capture.}
Direct-evidence probes are \probe{} actions aimed at evidence that bears on
task completion, so a higher rate means more specific checks. Aimless
scrolling and waiting are actions that make no progress toward such
evidence. Repeated targets count revisits to an already inspected target,
and dead-end probes count exploration that never reaches useful evidence.
Each proxy is computed independently over the same \probe{} interactions.

\begin{table}[H]
 \centering
 \caption{\method{}-4B directs more verification toward task-relevant
 evidence and reduces all three reported forms of unproductive exploration
 relative to prompted GPT-5.4. Rates are percentages, and $\Delta$ is
 \method{} minus GPT-5.4 in percentage points.}
 \label{tab:verification-proxies}
 \renewcommand{\arraystretch}{1.1}
 \begin{widetable}{\linewidth}{@{}lccc@{}}
 \toprule
 \textbf{Behavioral proxy} & \textbf{GPT-5.4} & \textbf{\method{}-4B} & \textbf{$\Delta$} \\
 \midrule
 Direct-evidence probes (\%) $\uparrow$ & 47.6 & \textbf{61.9} & $+14.3$ \\
 Aimless scroll and wait (\%) $\downarrow$ & 16.2 & \textbf{5.5} & $-10.7$ \\
 Repeated targets (\%) $\downarrow$ & 19.8 & \textbf{13.6} & $-6.2$ \\
 Dead-end probes (\%) $\downarrow$ & 25.4 & \textbf{21.7} & $-3.7$ \\
 \bottomrule
 \end{widetable}
\end{table}

\paragraph{More focused evidence acquisition.}
The trained verifier raises the direct-evidence rate by 14.3 points and
cuts aimless scrolling and waiting by 10.7 points. Repeated targets and
dead-end probes also decrease. Since the \probe{} is the same frozen model
in both cases, these differences come from the checks the \judge{} issues.

\section{Reward Reliability across \texorpdfstring{\probe{}}{Probe} Scales}
\label[appendix]{sec:probe-sensitivity}

Because the learned \judge{} decides what evidence to seek and the frozen
\probe{} only executes the request, the two can be scaled separately. We hold
the trained \judge{}, evaluation task set, and verification protocol fixed
and vary only the \probe{} within the Holo3.1 family, from 35B-A3B to 9B
and 4B. Each reported agreement aggregates
repeated evaluation trials across multiple random seeds.

\begin{table}[H]
    \centering
    \caption{Reward agreement remains stable across Holo3.1 \probe{} scales.
    The trained \judge{}, evaluation task set, and active-verification
    protocol are fixed, and only the frozen \probe{} changes. Confidence intervals
    are 95\% Wilson intervals.}
    \label{tab:probe-sensitivity}
    \renewcommand{\arraystretch}{1.1}
    \begin{widetable}{\linewidth}{@{}lcc@{}}
        \toprule
        \textbf{\probe{}} & \textbf{Agreement (\%)} & \textbf{95\% CI} \\
        \midrule
        Holo3.1-35B-A3B & $82.00$ & $[80.08, 83.78]$ \\
        Holo3.1-9B & $82.73$ & $[80.83, 84.47]$ \\
        Holo3.1-4B & $82.07$ & $[80.14, 83.84]$ \\
        \bottomrule
    \end{widetable}
\end{table}

\paragraph{Sensitivity to \probe{} scale.}
Agreement stays within $0.73$ points across the three \probe{} models. The
confidence intervals overlap closely, and agreement does not change
monotonically with model size. Pairwise two-sample proportion tests give $p=0.5843$ (35B-A3B
vs. 9B), $p=0.9639$ (35B-A3B vs. 4B), and $p=0.6158$ (9B vs. 4B). The 4B
\probe{} reaches $82.07\%$, against $82.00\%$ for 35B-A3B. Within the
Holo3.1 family, the \probe{} can therefore be reduced to 4B without
retraining the \judge{}.

\section{Evaluation Details}
\label[appendix]{sec:evaluation-details}

We use fixed inference settings for each component during reward evaluation.
The \judge{} uses a sampling temperature of $1.0$. The frozen \probe{} uses
a sampling temperature of $0.0$, a maximum output length of $8192$
tokens per API request, and at most $15$ interaction steps per probing round.
The maximum number of probing rounds follows the main experimental setup
(\cref{sec:experimental-setup}).

\paragraph{Trajectory screening and cost accounting.}
We re-execute each Actor trajectory six times from the clean task snapshot and
retain it only when all six re-executions yield the same executable outcome and
the mean pHash similarity over their final three screenshots exceeds $0.95$.
Verifier trials then run in separate fresh re-executions.
In cost accounting, trials that never call the \probe{} count as zero
tokens, and image tokens follow Qwen-VL accounting.
VAGEN has no separate Probe, so its full active loop is billed.
We measure cost in \probe{} tokens because this is the cost the
verification strategy controls. The \judge{} decides which checks the
\probe{} executes and when probing stops. Within each benchmark, every
prompted and trained \method{} \judge{} drives the same frozen \probe{}, so
these tokens are counted on one model with one tokenizer.
\judge{} tokens are not comparable in the same way. Tokenizers differ across
backbones, closed models such as GPT-5.4 report reasoning only as
provider-side counts, and per-token prices differ by orders of magnitude
between a locally served 1B--4B model and a proprietary API. Counting
\judge{} cost would therefore only widen the advantage of \method{} over
proprietary \judge{} models.

The \judge{} is paired with an environment-specific frozen \probe{}.
Within each benchmark, prompted and trained \method{} use the same \probe{}
configuration, so any difference between them comes from the learned
verification policy.
\Cref{tab:probe-evaluation-settings} summarizes the default configurations.

\begin{table}[H]
    \centering
    \caption{Default frozen \probe{} configurations used for reward evaluation.}
    \label{tab:probe-evaluation-settings}
    \renewcommand{\arraystretch}{1.1}
    \begin{widetable}{0.7\linewidth}{@{}ll@{}}
        \toprule
        \textbf{Benchmark} & \textbf{\probe{}} \\
        \midrule
        OSWorld & Holo3.1-35B-A3B \\
        MobileWorld & GUI-Owl-1.5-32B-Think \\
        WebArena-Verified & Holo3.1-35B-A3B \\
        \bottomrule
    \end{widetable}
\end{table}

\section{REINFORCE++ Optimization}
\label[appendix]{sec:reinforcepp-details}

We use the single-rollout REINFORCE++ estimator~\citep{reinforcepp}, as
implemented in Slime. It assigns the terminal outcome reward to the
generated \judge{} tokens in a complete verification rollout and normalizes
their advantages globally, as in \cref{eq:judge-whitening}. It therefore
needs neither a same-state rollout group nor a learned value model. The main
text states the underlying policy-gradient objective. Our implementation
uses the standard PPO-style clipped surrogate:
\begin{equation}
    \begin{aligned}
        \rho_{i,k,t}(\phi)
        &=
        \frac{\pi^J_{\phi}(y_{i,k,t}\mid h_{i,k,t})}
             {\pi^J_{\mathrm{old}}(y_{i,k,t}\mid h_{i,k,t})},
        \qquad
        \bar{\rho}_{i,k,t}(\phi)
        =
        \operatorname{clip}\!\left(
            \rho_{i,k,t}(\phi),
            1-\epsilon_{\mathrm{low}},
            1+\epsilon_{\mathrm{high}}
        \right), \\
        \mathcal{L}_{\mathrm{clip}}(\phi)
        &=
        -\frac{1}{|\mathcal{A}_{\mathcal{B}}|}
        \sum_{(i,k,t)\in\mathcal{A}_{\mathcal{B}}}
        \min\!\left\{
            \rho_{i,k,t}(\phi)A_{i,k,t},
            \bar{\rho}_{i,k,t}(\phi)A_{i,k,t}
        \right\}.
    \end{aligned}
    \label{eq:reinforcepp-clipped-loss}
\end{equation}

\section{Implementation Details}
\label[appendix]{sec:implementation-details}

\paragraph{Training implementation.}
We implement cold-start Judge SFT using LLaMA-Factory.
The Qwen3-VL-4B Judge is fine-tuned on eight GPUs with
DeepSpeed ZeRO-3. Online RL uses Slime with Megatron Core
for parameter updates and SGLang for rollout generation.
For online Judge RL, we allocate eight NVIDIA A800-SXM4
GPUs (80\,GB each) to parameter updates and eight additional
GPUs to rollout generation. Parameter updates use eight-way
tensor parallelism, and rollout generation uses single-GPU
engines. Training uses BF16 precision.
Objectives and hyperparameters are given in
\cref{sec:reinforcepp-details,sec:lantern-training-details}.

\paragraph{Inference implementation.}
We also run \judge{} and \probe{} inference on Huawei
Ascend 910B3 accelerators. The eight-NPU configuration
comprises three \probe{} instances using two NPUs each
and two \judge{} instances using one NPU each.
Inference uses BF16 precision with vLLM 0.18.0 and
vLLM-Ascend 0.18.0, running on CANN 9.0.0,
PyTorch 2.9.0, and \texttt{torch-npu} 2.9.0.post2.
Model inference and GUI execution run in separate
processes and communicate through API requests.
Model choices and decoding settings are given in
\cref{sec:evaluation-details}.

\paragraph{Desktop environments.}
The Ascend-based desktop evaluation uses 16 independent
x86 Ubuntu 22.04.3 virtual machines at a resolution of
$1920\times1080$, each with four virtual CPUs and
4\,GiB of memory. The virtual machines run under
QEMU 8.2.10 with TCG emulation on ARM64 hosts.
Environment initialization, trajectory re-execution,
and verifier isolation follow the protocol in
\cref{sec:evaluation-details}.

\section{\method{} Training Details}
\label[appendix]{sec:lantern-training-details}

\paragraph{Cold-start \judge{} SFT.}
We use the same cold-start SFT procedure for the Qwen3-VL-2B-Thinking,
Qwen3-VL-4B-Thinking, and MiniCPM-V-4.6-Thinking \judge{} backbones.
Each backbone is fully SFT-tuned on approximately $1{,}270$ Kimi-K2.6
verification traces retained by the rejection-sampling procedure in
\cref{sec:sft-coldstart}. These traces are generated over OSWorld trajectories from EvoCUA-32B, Holo3.1-35B-A3B, Qwen3-VL-4B-Thinking, and Qwen3.6-27B. The vision tower and multimodal projector remain
frozen while the language model is updated. Previous assistant turns serve
only as context, and the loss is applied to the current assistant turn.
Thinking content from earlier turns is removed during data construction.

\begin{table}[H]
    \centering
    \caption{Shared cold-start SFT configuration across \judge{} backbones.}
    \label{tab:judge-sft-hyperparameters}
    \renewcommand{\arraystretch}{1.1}
    \begin{widetable}{\linewidth}{@{}l>{\raggedright\arraybackslash}p{0.66\linewidth}@{}}
        \toprule
        \textbf{Setting} & \textbf{Value} \\
        \midrule
        \judge{} backbones & Qwen3-VL-2B-Thinking; Qwen3-VL-4B-Thinking; MiniCPM-V-4.6-Thinking \\
        Teacher supervision & Kimi-K2.6 Judge traces \\
        Training examples & Approximately $1{,}270$ \\
        Random seed & $1979$ \\
        Fine-tuning scope & Full SFT; frozen vision tower and multimodal projector; trainable language model \\
        Target masking & Current assistant turn only; previous turns used as context \\
        Effective global batch size & $16$ \\
        Optimization & Learning rate $3\times10^{-6}$, cosine schedule, weight decay $0.01$ \\
        \bottomrule
    \end{widetable}
\end{table}

\paragraph{Online \method{} training.}
\cref{tab:lantern-hyperparameters} summarizes the training protocol shared
by the Qwen3-VL-2B-Thinking, Qwen3-VL-4B-Thinking, and
MiniCPM-V-4.6-Thinking \judge{} backbones.  The OSWorld training set, Actor
pool, \probe{}, verification budget, RL objective, and optimization
configuration are identical across backbones.

\begin{table}[H]
    \centering
    \caption{Shared hyperparameters for online \method{} training across
    \judge{} backbones.}
    \label{tab:lantern-hyperparameters}
    \renewcommand{\arraystretch}{1.1}
    \begin{widetable}{\linewidth}{@{}l>{\raggedright\arraybackslash}p{0.66\linewidth}@{}}
        \toprule
        \textbf{Setting} & \textbf{Value} \\
        \midrule
        Trainable \judge{} backbones & Qwen3-VL-2B-Thinking; Qwen3-VL-4B-Thinking; MiniCPM-V-4.6-Thinking \\
        Initialization & Rejection-sampled SFT \\
        Training set & OSWorld ($280$ tasks) \\
        Fixed Actor pool & EvoCUA-32B; Qwen3-VL-4B-Thinking; Holo3.1-35B-A3B; Qwen3.6-27B \\
        Fixed \probe{} & Holo3.1-35B-A3B \\
        Verification budget & At most $K{=}2$ \probe{} rounds and $15$ GUI steps per round \\
        Advantage estimator & REINFORCE++ with batch normalization \\
        Optimization & Adam, learning rate $1\times10^{-6}$, weight decay $0.1$ \\
        \bottomrule
    \end{widetable}
\end{table}

\paragraph{Downstream Actor RL.}
All downstream reward-source comparisons start from the same
Qwen3-VL-4B-Thinking Actor and use the same training tasks, rollout budget,
GRPO objective, and optimization configuration. Only the reward provider
changes. When \method{} supplies the rewards, both the learned \judge{} and
the Holo3.1 \probe{} are frozen. \Cref{tab:policy-rl-hyperparameters} lists
the shared settings.

\begin{table}[H]
    \centering
    \caption{Shared hyperparameters for downstream GUI Actor RL across reward
    sources.}
    \label{tab:policy-rl-hyperparameters}
    \renewcommand{\arraystretch}{1.1}
    \begin{widetable}{\linewidth}{@{}l>{\raggedright\arraybackslash}p{0.66\linewidth}@{}}
        \toprule
        \textbf{Setting} & \textbf{Value} \\
        \midrule
        Actor model & Qwen3-VL-4B-Thinking \\
        Training set & OSWorld ($206$ single-application tasks) \\
        Interaction & Screenshot observations, PyAutoGUI actions, $30$-step horizon \\
        GRPO batch & $8$ tasks $\times$ $8$ rollouts \\
        Advantage estimator & GRPO \\
        PPO clipping & $(0.20, 0.28)$ \\
        Regularization & KL $0.01$ (low-variance); entropy $0$ \\
        Optimization & Adam, learning rate $1\times10^{-6}$, weight decay $0.1$ \\
        \bottomrule
    \end{widetable}
\end{table}
\FloatBarrier

\paragraph{Terminal reward.}
\label[appendix]{sec:verification-reward-details}
\probe{} usage is binary: $p_i=1$ if rollout $z_i$ contains at least one \probe{} invocation, regardless of the number of \judge{}--\probe{} rounds.
All rewards use the executable outcome $Y_i$ recorded before \probe{} interaction. The exact schedule is given in \cref{tab:verification-reward-schedule}.

\begin{table}[H]
    \centering
    \caption{Terminal reward schedule used for online \judge{} RL.}
    \label{tab:verification-reward-schedule}
    \renewcommand{\arraystretch}{1.1}
    \begin{widetable}{0.7\linewidth}{@{}lcc@{}}
        \toprule
        \textbf{Outcome} & \textbf{\probe{} used} & \textbf{Reward} \\
        \midrule
        Correct & No  & $+1.0$ \\
        Correct & Yes & $+0.7$ \\
        Incorrect & No  & $-0.7$ \\
        Incorrect & Yes & $-0.3$ \\
        Unresolved or invalid & -- & $-1.0$ \\
        \bottomrule
    \end{widetable}
\end{table}

An output counts as unresolved or invalid if the final decision is \texttt{uncertain}, the output cannot be parsed, or the executable evaluator returns no metadata. Under this schedule, a correct verdict is worth more without probing, while a wrong verdict costs less after probing, so the reward favors probing exactly when the screenshot-only verdict is likely to be wrong.

\section{Verification Behavior during Downstream Actor RL}
\label[appendix]{sec:behavior-analysis}

\begin{figure}[H]
    \centering
    \includegraphics[width=\linewidth]{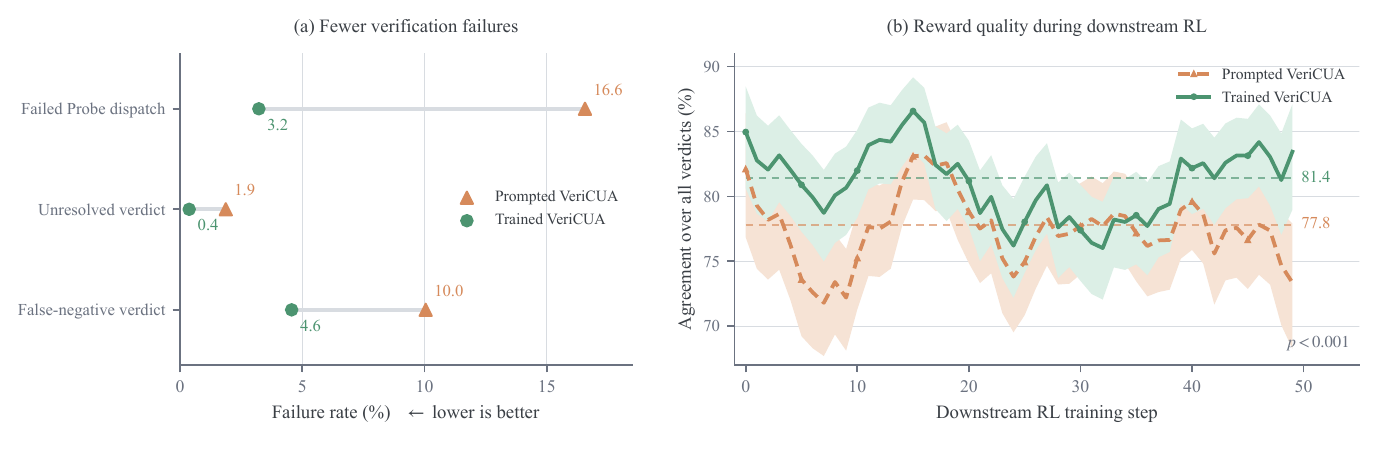}
    \caption{Learned verification behavior during downstream Actor RL. \textbf{(a)} Training reduces failed \probe{} dispatches, unresolved verdicts, and false-negative verdicts. \textbf{(b)} The learned verifier maintains higher agreement as the Actor evolves. Unresolved outputs count as incorrect in \textbf{(b)}. Shaded bands show 95\% Wilson confidence intervals.}
    \label{fig:behavior-analysis}
\end{figure}

We log every \judge{} decision from two downstream Actor-RL runs, one rewarded by prompted \method{} and the other by trained \method{}. The runs share the initial Actor, tasks, rollout budget, and optimization configuration. Comparing their decisions with executable outcomes shows how verification behaves as the Actor changes. \Cref{tab:behavior-metrics,fig:behavior-analysis} report how often the \judge{} probes, how reliably its \probe{} requests are dispatched, and how often its verdicts match the executable outcome.

\begin{table}[H]
    \centering
    \caption{Training reduces verification failures and improves terminal reward accuracy during downstream Actor RL.}
    \label{tab:behavior-metrics}
    \renewcommand{\arraystretch}{1.1}
    \begin{widetable}{\linewidth}{@{}lccc@{}}
        \toprule
        \textbf{Behavioral metric} & \textbf{Prompted} & \textbf{Trained} & \textbf{$p$} \\
        \midrule
        \probe{} attempt rate (\%) $\downarrow$ & 5.70 & \textbf{4.28} & $0.013$ \\
        Failed \probe{} dispatch (\%) $\downarrow$ & 16.6 & \textbf{3.2} & $3\times10^{-4}$ \\
        Unresolved uncertain (\%) $\downarrow$ & 1.9 & \textbf{0.4} & --- \\
        False-negative rate (\%) $\downarrow$ & 10.0 & \textbf{4.6} & $1\times10^{-7}$ \\
        Agreement with executable evaluator (\%) $\uparrow$ & 77.8 & \textbf{81.4} & $<0.001$ \\
        \bottomrule
    \end{widetable}
\end{table}

\paragraph{Dispatch failures and unresolved verdicts.}
Failed \probe{} dispatch falls from $16.6\%$ to $3.2\%$, and unresolved \texttt{uncertain} outputs fall from $1.9\%$ to $0.4\%$ (\cref{fig:behavior-analysis}(a)). The \probe{} is frozen in both runs, so the improvement comes from the \judge{}. It learns to issue checks that the \probe{} can act on and to reach a terminal decision from the returned evidence. It does so with $25\%$ fewer \probe{} attempts (\cref{tab:behavior-metrics}).

\paragraph{Agreement over the course of Actor training.}
\method{} leads prompted \method{} in agreement by $4.2$ points in the first half of training and $3.2$ points in the second (\cref{fig:behavior-analysis}(b)). Its false-negative rate is also lower ($4.6\%$ vs.\ $10.0\%$), so fewer successful Actor rollouts receive a zero reward.

\section{Prompt Templates}
\label[appendix]{sec:prompt-templates}

For reproducibility, we provide the prompts used by \method{} and its prompted
variants, covering verification-check generation and verdict prediction by the \judge{},
environment-specific evidence acquisition by the \probe{}, and the few-shot
criteria references used by prompted \method{} configurations. Card colors follow the roles
in the method overview figure: \textcolor{RAJudge}{\textbf{green}} for the
\judge{}, \textcolor{RAProbe}{\textbf{blue}} for the \probe{}, and
\textcolor{RARef}{\textbf{warm gold}} for shared few-shot references.

\subsection{\texorpdfstring{\judge{}}{Judge} Prompts}

\begin{JudgePrompt}{System prompt}
\begin{PromptBlock}
You are an expert in evaluating the performance of a GUI agent. The agent is designed to help a human user complete tasks across desktop applications and web pages.
Your job is to:
1. write concise, concrete, and verifiable task-success criteria;
2. judge whether each criterion is currently success, failure, or uncertain based on the provided evidence;
3. decide the overall task status: success, failure, or uncertain.

There are several types of tasks (not limited to the list below):
1. Information seeking: The user wants to obtain information from a webpage, document, or application. The final response must contain the requested information or explicitly state that it is unavailable. If the agent returns an exception or irrelevant error content, the task is a failure.
2. Navigation: The user wants to navigate to a specific page, view, folder, screen, or application state. Examine the action history and final visual state. The agent's response is usually not required.
3. Content modification: The user wants to modify content in a webpage, document, file, setting, or application. Verify that the final state reflects the requested change.
4. File operations: The user wants to create, rename, move, copy, delete, compress, extract, or inspect files or folders. Verify that the filesystem state matches the request.
5. Application setup or configuration: The user wants to install, uninstall, enable, disable, or configure software, extensions, plugins, or settings. Verify that the final applied state indicates the requested setup/configuration is complete.
6. Media actions: The user wants to open, play, pause, record, view, or manipulate images, audio, or video. Verify that the media state reflects the requested action.
7. Communication: The user wants to read, compose, send, or retrieve emails/messages. Verify the intended content, recipient, and sent/retrieved state when applicable.
8. Terminal or command execution: The user wants to run shell commands. Verify that the commands were executed correctly and that outputs or side effects match the intent.

==Overall Status (check in this order)==
1. success: all criteria are success.
2. failure: otherwise, if any criterion is failure (or contradicted / done incorrectly).
3. uncertain: otherwise (no failure, but at least one uncertain).

==Output exactly this format==
<criteria>
1. <concrete task-success standard>:<success|failure|uncertain>
2. <concrete task-success standard>:<success|failure|uncertain>
3. <concrete task-success standard>:<success|failure|uncertain>

</criteria>
Status: <success|failure|uncertain>

==Example: Open report.docx in LibreOffice Writer, change the document title to "2026 Q1 Sales Review", and save the file.==
<criteria>
1. report.docx is opened in LibreOffice Writer:success
2. The document title text reads exactly "2026 Q1 Sales Review":uncertain
3. The change is persisted to report.docx (file saved, no unsaved-changes indicator):failure
</criteria>
Status: failure

==Keep it in mind==
1. If previous criteria are not provided, produce 3 to 5 criteria.
2. If previous criteria are provided, keep the criteria text unchanged. The probe agent has already assessed the status of each criterion - treat the probe's assessment as your default baseline. Only override a status (e.g., downgrade 'success' back to 'uncertain') when you observe clear, direct visual contradiction: for example, a file that is visibly unmodified, a setting that is clearly still at its default value, or an error message that proves the task was not completed. Do not downgrade to 'uncertain' merely because the evidence is limited, indirect, or ambiguous.
3. Use only the user task as requirements; do not add unrelated or overly strict criteria.
4. Do not treat action-history labels like Done, Success, Completed, saved, clicked Save/Submit/OK/Apply as evidence by themselves.
\end{PromptBlock}
\end{JudgePrompt}

\begin{JudgePrompt}{First-round input}
\begin{PromptBlock}
User Intent:
{task_instruction}

Criteria:
No criteria available.

Action History:
Step 1: {actor_action_1}
Step 2: {actor_action_2}
...
Step N: {actor_action_N}

The following {number_of_screenshots} screenshots are the latest observations,
shown in chronological order.

[image: latest trajectory screenshot 1]
[image: latest trajectory screenshot 2]
...
[image: latest trajectory screenshot history_n]
\end{PromptBlock}
\end{JudgePrompt}

\begin{JudgePrompt}{Input after probing}
\begin{PromptBlock}
User Intent:
{task_instruction}

Criteria:
{previous_criteria_with_statuses}

The attached screenshot(s) are evidence collected by the probe for the
criteria, shown in chronological order.

Evidence image {criterion_number} for the current criteria.
[image: screenshot collected while checking criterion {criterion_number}]
\end{PromptBlock}
\end{JudgePrompt}

If no check-associated screenshot is returned:

\begin{JudgePrompt}{Input after probing (no evidence returned)}
\begin{PromptBlock}
User Intent:
{task_instruction}

Criteria:
{previous_criteria_with_statuses}

The probe agent did not find corresponding evidence for the criteria. Please
judge based on the following {number_of_screenshots} latest screenshots from
the original agent trajectory, shown in chronological order.

[image: latest trajectory screenshots]
\end{PromptBlock}
\end{JudgePrompt}

\subsection{Environment-Specific \texorpdfstring{\probe{}}{Probe} Prompts}

\begin{ProbePrompt}{Desktop system prompt}
\begin{PromptBlock}
You are a GUI evidence probe running inside a desktop GUI agent loop.

A separate GUI agent {agent_name} has already attempted the user's task in this environment. Your task is not to redo the whole task; your task is to inspect the current GUI state, collect just enough evidence, and update the judge criteria using the `criteria` tool.

You operate an {platform} desktop with internet access.
The current date is {current_date}.
The home directory is "{home_dir}".
The computer password is "{client_password}" when sudo is needed.

At each step, choose exactly one tool call from two modes:
1. Exploration: use action tools (click, write, keypress, scroll, wait) to navigate and reveal evidence.
2. Criteria update: use the `criteria` tool to formally update a criterion's state.

IMPORTANT RULES about the `criteria` tool:
- You MUST call the `criteria` tool to update each criterion when you observe evidence. Reasoning about evidence in your thought/note is NOT sufficient - the judge only sees criteria updates made through the `criteria` tool call.
- Update criteria one at a time. Each `criteria` call updates exactly one criterion number.
- Only call `answer` with "DONE" AFTER you have called `criteria` for every uncertain criterion that you have evidence for. If you have not called `criteria` at all, do NOT call `answer`.

Follow the screenshot in each <observation>. Choose exactly one tool call per step.
Coordinates must be integers in [0, 1000] relative to the exact screenshot you see, with origin at the top-left.

Use write only after the target input is focused. Use click first if focus is uncertain. Prefer keypress for shortcuts and Enter. Scroll when content may be off-screen.

Do not perform actions that modify the environment (e.g. saving files, deleting, editing). Your role is to observe and verify, not to change the state.

<output_format>
{probe_json_schema}
</output_format>
\end{PromptBlock}
\end{ProbePrompt}

\begin{ProbePrompt}{Desktop step input}
\begin{PromptBlock}
Generate the next step: either explore with an action tool, or update exactly
one criterion with the `criteria` tool when the current screenshot provides
evidence.

Current Criteria:
{current_criteria}

Previous actions:
{previous_probe_actions_or_None}

<observation>
[image: current GUI screenshot]
</observation>
\end{PromptBlock}
\end{ProbePrompt}

\begin{ProbePrompt}{Mobile system prompt}
\begin{PromptBlock}
You are an evidence probe inspecting the current state of an Android device
after another agent attempted a task. Do not repeat the whole task and do not
make destructive or irreversible changes. Use the native mobile actions only
to reveal enough on-screen evidence for the judge. Prefer navigation, opening
relevant views, scrolling, and wait. Do not ask the user, call MCP tools,
submit answers, or modify user data. When sufficient evidence is visible,
terminate.

Original task:
{task_instruction}

Current judge criteria:
{current_criteria}
\end{PromptBlock}
\end{ProbePrompt}

\begin{ProbePrompt}{Mobile step input}
\begin{PromptBlock}
Native-agent observation:
  screenshot: [current Android screenshot]
  tool_call: {previous environment tool call, if any}
  ask_user_response: {environment response, if any}
\end{PromptBlock}
\end{ProbePrompt}

\begin{ProbePrompt}{Web system prompt}
\begin{PromptBlock}
You are a visual evidence probe inspecting a live web browser after another agent attempted a task. Do not redo the task. Reveal only enough state to judge the criteria, using low-risk browser navigation. Never create, edit, delete, send, submit, purchase, or otherwise mutate data.
Stay inside the currently configured WebArena sites and the current site origin. Never follow or enter a public-web URL such as Google, OpenStreetMap, or another external domain. If useful evidence is not visible, use scroll, go_back, or criteria/answer instead of leaving WebArena.

Choose exactly one structured tool call per turn. Coordinates are integers in [0, 1000] relative to the screenshot. Use `criteria` to update exactly one numbered criterion when visible evidence supports it. The judge only receives evidence saved through `criteria`. Use `answer` with DONE once useful inspection is complete.

Original task:
{instruction}

Current criteria:
{criteria}

Current date: {current_date}

Output must match this JSON schema:
{schema}
\end{PromptBlock}
\end{ProbePrompt}

\begin{ProbePrompt}{Web step input}
\begin{PromptBlock}
<observation>
Current URL: {current_url}
Open tabs: {open_pages_urls}
[image: current browser screenshot]
</observation>
\end{PromptBlock}
\end{ProbePrompt}

\subsection{Few-Shot Criteria References}
\label[appendix]{sec:fewshot-prompt-templates}

Few-shot prompted \method{} configurations receive two representative
references per benchmark, fixed across all runs. Each reference illustrates the
criteria format on an example task: the criteria only name the observable
final-state conditions that a task implies, and carry no verdict, so the
examples supply formatting guidance rather than task-specific evidence. The
same two references are prepended to every run of every few-shot
configuration compared in \cref{tab:main-consistency}.

Few-shot prompted \method{} configurations prepend two text-only examples of observable final-state
checks using the following shared prefix:

\begin{RefPrompt}{Shared few-shot prefix}
\begin{PromptBlock}
Reference examples below demonstrate how to derive concise, concrete, and
verifiable criteria from a user task. Their criteria contain no verdicts. The
examples are unrelated to the current task and are not evidence for it; do not
infer their outcomes.
\end{PromptBlock}
\end{RefPrompt}

\begin{RefPrompt}{OSWorld references}
\begin{PromptBlock}
==Reference example 1==
User Intent:
Go to the second slide and name its title as "Online Shopping" with the same
color, position and font size as the previous title.

Reference Criteria:
1. The second slide is active and its title reads "Online Shopping"
2. The title color matches the color of the previous slide's title
3. The title position matches the position of the previous slide's title
4. The title font size matches the font size of the previous slide's title

==Reference example 2==
User Intent:
Tomorrow, I'm scheduled to deliver a talk, and my PowerPoint slides and
speaking notes are saved on the desktop. Help me insert my planned remarks for
each slide into the "note" section of the PowerPoint as a reminder. I've
completed this task for some slides; assist me in completing the remaining part.

Reference Criteria:
1. The PowerPoint presentation file is opened in LibreOffice Impress
2. Slide 7's note section contains the text "You should also try to answer!"
3. The file has been saved without unsaved-changes indicators
4. All slides in the presentation have planned remarks inserted into their note sections
\end{PromptBlock}
\end{RefPrompt}

\begin{RefPrompt}{MobileWorld references}
{\color{RAPromptText}\normalsize\ttfamily\raggedright\setlength{\parindent}{0pt}
==Reference example 1==\\
User Intent:\\
{\normalfont\rmfamily\normalsize
请帮我算一下在淘店上最近1个月我总共花了多少钱。请只回答一个整数，不要返回任何其他文本。}\\[0.6em]
Reference Criteria:\\
1. The agent accessed the user's order history or spending records on Taobao.\\
2. The agent calculated the total spending for the last month and determined it to be 1196.\\
3. The agent's final response is a single integer.\\
4. No extraneous text or errors are returned in the response.\\[0.6em]
==Reference example 2==\\
User Intent:\\
In Mastodon, remove the posts with \#pets tag from bookmarks on my account.\\[0.6em]
Reference Criteria:\\
1. The agent navigated to the account's profile bookmarks section.\\
2. All bookmarked posts containing the \#pets tag have been removed.\\
3. The bookmark list no longer contains \#pets tagged content.\par}
\end{RefPrompt}

\begin{RefPrompt}{WebArena-Verified references}
\begin{PromptBlock}
==Reference example 1==
User Intent:
Delete all pending reviews with less than 4 stars

Reference Criteria:
1. All pending reviews with less than 4 stars have been removed from the list.
2. The current "Pending Reviews" page shows no entries with a star rating below 4 stars.

==Reference example 2==
User Intent:
How many reviews did our shop receive in Apr 2023?

Reference Criteria:
1. The filter for April 2023 (04/01/2023 to 04/30/2023) is correctly applied and displayed in the UI.
2. The total number of reviews in the filtered results is 351 (as shown by "351 records found" in the UI).
\end{PromptBlock}
\end{RefPrompt}

\end{document}